\documentclass[11pt]{article}
\def\spacingset#1{\def\baselinestretch{#1}\small\normalsize}
\usepackage{stmaryrd}
\SetSymbolFont{stmry}{bold}{U}{stmry}{m}{n}
\usepackage{mathrsfs}
\usepackage{amsmath}
\usepackage{amssymb}
\usepackage{multirow}
\usepackage{cite}
\usepackage[dvips]{graphicx}

\usepackage{color}
\newcommand{\ba}{\begin{array}}
\newcommand{\ea}{\end{array}}
\newcommand{\be}{\begin{displaymath}}
\newcommand{\ee}{\end{displaymath}}
\newcommand{\ben}{\begin{equation}}
\newcommand{\een}{\end{equation}}
\newcommand{\bena}{\begin{eqnarray}}
\newcommand{\eena}{\end{eqnarray}}
\newcommand{\beqa}{\begin{eqnarray*}}
\newcommand{\enqa}{\end{eqnarray*}}

\newcommand{\bc}{\begin{center}}
\newcommand{\ec}{\end{center}}
\newcommand{\bi}{\begin{itemize}}
\newcommand{\ei}{\end{itemize}}
\newcommand{\benu}{\begin{enumerate}}
\newcommand{\eenu}{\end{enumerate}}
\newcommand{\bdes}{\begin{description}}
\newcommand{\edes}{\end{description}}
\newcommand{\bt}{\begin{tabular}}
\newcommand{\et}{\end{tabular}}

\newcommand{\circlambda}{\mbox{$\Lambda$
             \kern-.85em\raise1.5ex
             \hbox{$\scriptstyle{\circ}$}}\,}

\usepackage{hyperref}  

\usepackage{amsfonts}       % 提供数学黑体符号（如\mathbb{R}）  
\usepackage{amssymb}        % 扩展数学符号（如\varnothing, \triangle）
\usepackage{amsmath}        % 高级数学环境（cases, matrix, align等）
\usepackage{nicefrac}       % 紧凑分数（如 ½ 显示为`\nicefrac{1}{2}`）
\usepackage{microtype}      % microtypography
\usepackage{xcolor}         % colors
\usepackage{subcaption}    % for subfigure environment
\usepackage{siunitx}
\usepackage{booktabs}        % 专业表格线（\toprule, \midrule, \bottomrule）
\usepackage{multirow}        % 合并多行单元格（\multirow{2}{*}{文字}）
\usepackage{colortbl}        % 表格着色（与xcolor联用，\rowcolor）
\usepackage{adjustbox}       % 调整表格/图片大小（max width等）
\usepackage{graphicx}  % 核心图片插入支持
\usepackage{enumitem}
\usepackage{wrapfig} % 图文环绕支持
\usepackage{float}
\usepackage[ruled,boxed]{algorithm2e}
\makeatletter
\@fpsep\textheight
\makeatother

\newcommand{\qed}{\nobreak \ifvmode \relax \else
      \ifdim\lastskip<1.5em \hskip-\lastskip
      \hskip1.5em plus0em minus0.5em \fi \nobreak
      \vrule height0.75em width0.5em depth0.25em\fi}

\begin{document}

\title{\Large\bf ALISE: Annotation-Free LiDAR Instance Segmentation for Autonomous Driving}
\author{
Yongxuan Lyu
\thanks{
All authors are from Department of Electronic Engineering and Information Science, University of Science \& Technology of China 
}
\and Guangfeng Jiang
\and Hongsi Liu
\and Jun Liu
\thanks{e-mail:junliu@ustc.edu.cn}
}

% \author{
% Jun Liu
% \thanks{J. Liu, and H. Li are with
% the Department of Electrical and Computer Engineering, Stevens
% Institute of Technology, Hoboken, NJ 07030 USA. (e-mail:
% jun\_liu\_math@hotmail.com, hongbin.li@stevens.edu)}
% \and Hongbin Li$^{\dagger}$
% \and Braham~Himed
% \thanks{B. Himed is with AFRL/RYMD, 2241 Avionics Circle,
% Bldg 620, Dayton, OH45433.
% (e-mail: braham.himed@wpafb.af.mil)}
% }
\maketitle
\spacingset{1.5}

\begin{center}
\section*{Abstract}
\end{center}
The manual annotation of outdoor LiDAR point clouds for instance segmentation is extremely costly and time-consuming. Current methods attempt to reduce this burden but still rely on some form of human labeling. To completely eliminate this dependency, we introduce ALISE, a novel framework that performs LiDAR instance segmentation without any annotations. The central challenge is to generate high-quality pseudo-labels in a fully unsupervised manner. Our approach starts by employing Vision Foundation Models (VFMs), guided by text and images, to produce initial pseudo-labels. We then refine these labels through a dedicated spatio-temporal voting module, which combines 2D and 3D semantics for both offline and online optimization. To achieve superior feature learning, we further introduce two forms of semantic supervision: a set of 2D prior-based losses that inject visual knowledge into the 3D network, and a novel prototype-based contrastive loss that builds a discriminative feature space by exploiting 3D semantic consistency. This comprehensive design results in significant performance gains, establishing a new state-of-the-art for unsupervised 3D instance segmentation. Remarkably, our approach even outperforms MWSIS, a method that operates with supervision from ground-truth (GT) 2D bounding boxes by a margin of 2.53\% in mAP (50.95\% vs. 48.42\%).

\textbf{Keywords} --
Label-free learning, 3D instance segmentation, multi-modal, autonomous driving.


\newpage
\section{Introduction}

3D point cloud segmentation tasks constitute a fundamental research area in computer vision. Recently, impressive advancements in LiDAR point cloud segmentation have been achieved, largely driven by the availability of high-quality autonomous driving datasets~\cite{sun2020scalability, caesar2020nuscenes, behley2019semantickitti, geiger2013vision} and advancements in network architectures~\cite{shi2020points, zhou2021cylinder3d, yan2018second, qi2017pointnet, zhao2021ptv1, wu2022ptv2, wu2024ptv3}. However, these tasks typically depend on dense point-wise annotations, the acquisition of which is labor-intensive and expensive. As such, lessening the need for such extensive manual labeling has substantial practical value.

Although prior works have investigated weakly-supervised (e.g., sparse point-level~\cite{chen2024foundation, liu2021one}, scribble-level~\cite{unal2022scribble}, and box-level labels~\cite{chibane2022box2mask, jiang2024mwsis}) and unsupervised methodologies~\cite{liu2023segment, chen2023towards, wang2024uniplv}, their primary focus has largely remained on semantic segmentation. However, instance segmentation presents a more formidable challenge as it requires distinguishing instances within the same semantic category.

\begin{figure*}[ht]
    \centering
    % \vspace{-3mm} % 控制与上方段落距离
    \includegraphics[width=0.7\textwidth]{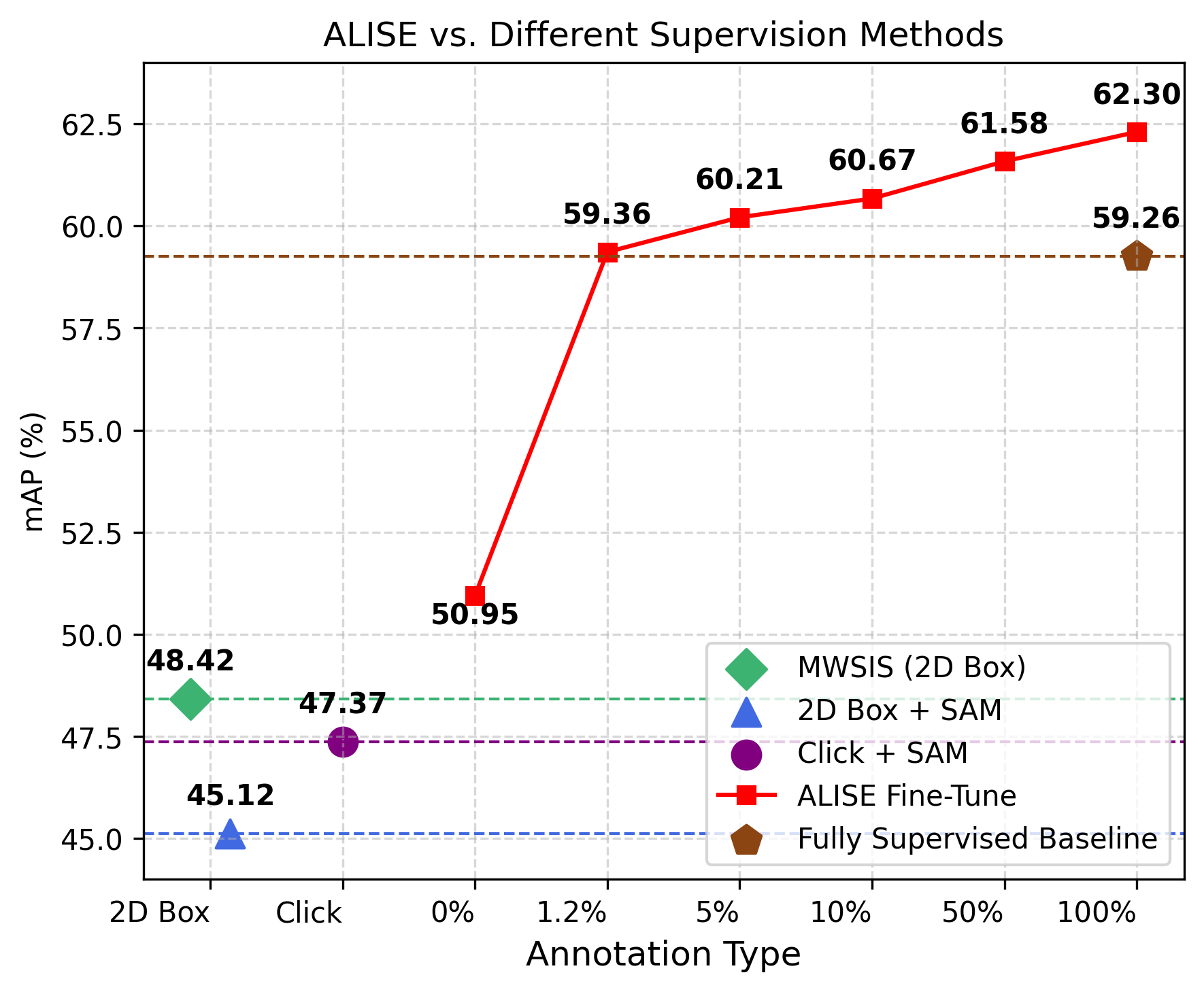}
    \caption{Performance comparison of ALISE against methods with different supervision types. Our label-free method ALISE (at 0\% GT) surpasses weakly supervised baselines. When fine-tuned with a small amount of GT labels, ALISE consistently outperforms the fully supervised baseline.}
    \label{fig:compare}
    % \vspace{1mm}
\end{figure*}

For 3D instance segmentation tasks, while certain weakly-supervised approaches have demonstrated encouraging outcomes, they still rely on some form of annotation. For instance, MWSIS~\cite{jiang2024mwsis} investigated weakly-supervised instance segmentation of outdoor LiDAR point clouds utilizing low-cost 2D bounding boxes as supervisory signals. Motivated by these advancements, we aim to develop a completely label-free framework that closes the gap with fully-supervised methods.

The powerful generalization capabilities of VFMs offer a promising avenue to generate 3D pseudo-labels from images, thereby eliminating the reliance on any manual annotation. However, this cross-modal transfer is fraught with challenges. VFMs can inevitably produce erroneous predictions, and pixel-to-point projection errors further introduce significant noise into the generated pseudo-labels.
To address these challenges, we propose a novel annotation-free 3D instance segmentation framework called ALISE, designed to fully leverage information from VFMs and robustly refine their noisy pseudo-labels. Firstly, we introduce an Unsupervised Pseudo-label Generation (UPG) module. Unlike prior works that directly generate one-hot labels, our UPG module preserves the VFM based semantic distribution across all classes. We then propose an Offline Refinement (OFR) strategy that generates pseudo-labels with higher quality by aggregating semantic priors from multiple adjacent frames for voxel-based semantic voting. 
Secondly, to fully exploit the image-based information, we design a VFM Priors-based Distillation (VPD) module to transfer rich knowledge to the 3D segmentation network. In addition, we introduce an Online Refinement (ONR) strategy during the training stage, which uses the network's own reliable predictions to correct noisy labels. Finally, we propose a Prototype-based Contrastive Learning (PCL) module to learn discriminative feature representations using dynamically updated prototypes. Our method achieves competitive performance for instance segmentation on both Waymo~\cite{sun2020scalability} and nuScenes~\cite{caesar2020nuscenes} datasets. It not only surpasses a wide range of weakly supervised approaches, but also exhibits impressive fine-tuning performance, exceeding the fully supervised baseline using merely 1.2\% of ground-truth annotations.
The main contributions of our work are summarized as follows:\begin{itemize}[nosep]
    \item We propose ALISE, a novel annotation-free framework for 3D instance segmentation that outperforms several weakly-supervised methods.

    \item We introduce a comprehensive pseudo-label generation and refinement pipeline. This includes: a UPG module that preserves the semantic distribution from VFMs, and a powerful tempoarl-based refinement strategy combining offline refinement (OFR) with online refinement (ONR).

    \item We design a multi-faceted supervision scheme, featuring a VPD module which distills the rich semantic knowledge of VFMs into the 3D segmentation network, and a PCL module that builds a dynamic feature prototype bank to learn discriminative point-wise representations.
\end{itemize}
\section{Related Work}
\subsection{Weakly-Supervised 3D Instance Segmentation}
While fully-supervised point cloud segmentation has progressed significantly, the associated dense annotation is prohibitively expensive. To mitigate this burden, a variety of weakly-supervised methods have been proposed. Some works utilize bounding boxes as supervision. For instance, Box2Mask~\cite{chibane2022box2mask} pioneered the use of 3D boxes. To further reduce annotation costs, MWSIS~\cite{jiang2024mwsis} successfully employed 2D bounding boxes for outdoor scenes. Another popular form of weak supervision involves using sparse clicks or scribbles~\cite{liu2021one, unal2022scribble}. YoCo~\cite{jiang2025you} first employed click annotation to outdoor 3d instance segmentation, which require significantly less annotation effort than boxes.
Despite their success in reducing the annotation workload, all these weakly-supervised approaches still rely on some form of manual labeling. In contrast, our work takes a leap forward by proposing a framework that operates in a completely annotation-free manner, entirely eliminating the need for human intervention in the labeling process.
\subsection{Label-Free 3D Segmentation}
Leveraging the remarkable performance of VFMs, several recent works have explored using image data to provide supervisory information for 3D segmentation. Some works like~\cite{liu2023segment} utilize contrastive learning to distill knowledge from powerful image-based models into 3D segmentation networks. Other works utilize the CLIP model~\cite{radford2021learning} to transfer open-vocabulary knowledge from 2D to 3D.
Methods such as OpenScene~\cite{peng2023openscene} and CLIP2Scene~\cite{chen2023clip2scene} project multi-view image features onto point clouds and distill semantic representations into 3D backbones, enabling zero-shot 3D segmentation without any 3D annotations.
UniPLV~\cite{wang2024uniplv} further bridges images and point clouds through intermediate text embeddings, while SAL~\cite{ovsep2024better} predicts CLIP-aligned tokens for point cloud segments, which are directly matched to text embeddings for segment-level zero-shot inference. 
Other approaches~\cite{ma2024zopp} directly employ VFMs like GroundingDINO~\cite{liu2024grounding} and SAM~\cite{kirillov2023segment} to generate 3D pseudo-labels for network supervision.

However, these methods typically generate one-hot semantic labels, which can introduce noisy supervision when the VFM produces wrong predictions. To address this challenge, our UPG module preserves the VFM's predicted semantic distributions rather than collapsing them into one-hot labels. Subsequently, our OFR module refines the semantic labels of the current frame by aggregating semantic priors from multiple frames for voxel-based voting. 
Unlike prior work that relies on hard labels, our VPD module utilizes VFM's predicted smoothed probability distributions as a robust form of soft-label supervision. Furthermore, the PCL module enhances the network's semantic discrimination capabilities by selecting high-confidence predictions to update class-specific prototypes and then enforcing feature consistency through contrastive learning.

\section{Proposed Method}
% Our proposed ALISE achieves annotation-free 3D instance segmentation through a synergistic pipeline of pseudo-label generation, refinement, and model supervision. We first introduce the UPG module, which leverages VFMs to generate coarse 3D pseudo-labels. To enhance label quality, we employ a temporal pseudo-label refinement strategy, consisting of an offline step using VFM priors and an online step with teacher network predictions. The 3D segmentation network is then trained with two distinct supervision modules. The VPD distills rich semantic and feature knowledge from the VFM. The PCL module learns discriminative features by enforcing consistency with high-confidence prototypes. The overall architecture is illustrated in Fig.~\ref{fig:UPG} and Fig.~\ref{fig:train}.

\begin{figure*}[t]
    \centering
    % 将宽度设置为 \textwidth (整个文本区域的宽度)
    \includegraphics[width=0.95\textwidth]{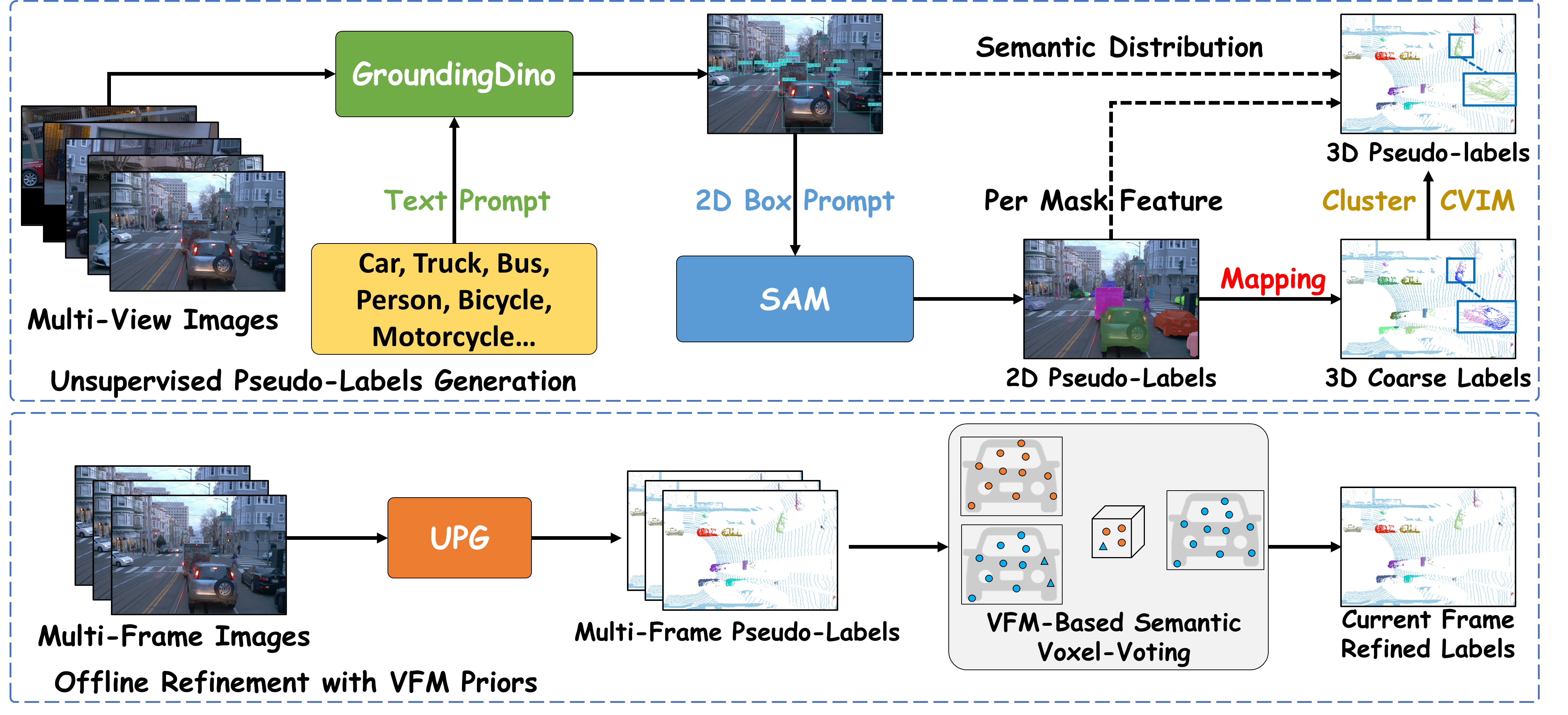}
    \vspace{-2mm}
    \caption{Illustration of the UPG module and the OFR module. Blue points represent the current frame, while orange points represent the adjacent frame. Different classes are indicated by using circles and triangles.}
    \label{fig:UPG}
    \vspace{-4mm}
\end{figure*}

\begin{figure*}[t]
    \centering
    % 将宽度设置为 \textwidth (整个文本区域的宽度)
    \includegraphics[width=0.95\textwidth]{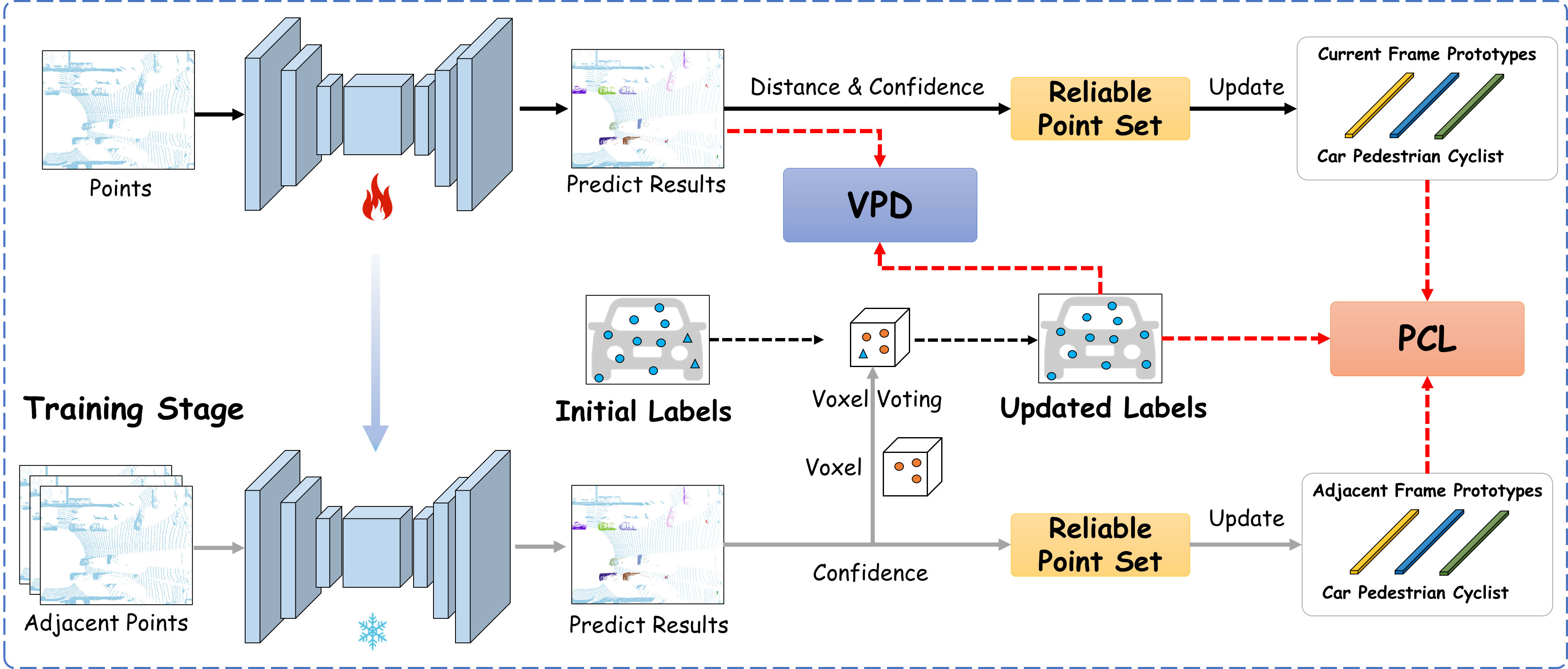}
    \caption{Illustration of the ONR module, the VPD module, and the PCL module in training stage.}
    \label{fig:train}
    \vspace{-4mm}
\end{figure*}

Our proposed ALISE framework achieves annotation-free 3D instance segmentation through a synergistic combination of modules for pseudo-label generation, refinement, and network training. Specifically, we first introduce the UPG module, which leverages VFMs to generate initial 3D pseudo-labels while preserving their full semantic distributions. 
To enhance the quality of these initial labels, we devise a two-stage refinement process. First, an OFR strategy improves the labels by aggregating semantic information from adjacent frames. Subsequently, during the training loop, an ONR strategy further updates the labels using predictions from a teacher network. 
The 3D segmentation network is trained under a multi-faceted supervision scheme. The VPD module transfers rich semantic knowledge from the 2D domain, while the PCL module facilitates the learning of discriminative point-wise representations. The overall architecture is illustrated in Fig.~\ref{fig:UPG} and Fig.~\ref{fig:train}.

\subsection{Multi-Modal Spatial Alignment}
\label{sec:multimodal_alignment}
To correlate 3D point clouds with 2D image information, we project each 3D point $p^{3d} = (x, y, z)$ onto the image plane to obtain its corresponding 2D pixel coordinates $p^{2d} = (u, v)$. This projection, denoted by the function $\pi: \mathbb{R}^3 \to \mathbb{R}^2$, is performed using the standard camera model transformation with known sensor calibration parameters:
\begin{equation}
\textstyle
z_c \cdot [u, v, 1]^T = \mathbf{K} \cdot \mathbf{T} \cdot [x, y, z, 1]^T,
\label{eq:projection_formula}
\end{equation}
where $z_c$ is the point's depth in the camera coordinate system, $\mathbf{K}$ is the camera intrinsic matrix, and $\mathbf{T}$ is the extrinsic transformation matrix from LiDAR to camera coordinates.

\subsection{Unsupervised Pseudo-Label Generation}
\label{sec:UGP}
Our framework begins by generating initial 3D pseudo-labels from multi-view images using a pipeline of VFMs. This process involves three main steps: 2D open-vocabulary detection, mask generation, and 3D label generation and refinement.

\noindent\textbf{2D Detection and Confidence Estimation.}
We first employ GroundingDINO~\cite{liu2024grounding} to perform open-vocabulary 2D detection using text prompts for our target classes. A special merging strategy is applied to composite objects like cyclists by associating each detected bicycle with a nearby person whose bounding box lies above the bicycle and is horizontally close, as shown in Fig~\ref{fig:cyc}. If multiple candidates exist, the closest one is selected. For each detected bounding box $b_i$, the model outputs a raw probability distribution across all text prompts. We process this to create a clean semantic distribution vector, $P_i^{\text{2D}}$, over our $C$ predefined classes by taking the maximum probability among all prompts associated with each class. The overall confidence of the detection $S_i$ is then defined as the maximum value within this vector:
\begin{equation}
\textstyle
    S_i = \max_{c=1 \dots C} P_i^{\text{2D}}(c)
\label{eq:instance_confidence}
\end{equation}

\noindent\textbf{3D Pseudo-Label Generation and Refinement.}
The detected 2D bounding boxes $b_i$ serve as prompts for the   SAM~\cite{kirillov2023segment}. From the three mask candidates generated by SAM for each prompt, we select the one with the highest predicted score, denoted as $m_i$. This 2D mask is then lifted to 3D by projecting the entire point cloud $\mathcal{P}$ onto the image plane and selecting all points whose projections $\pi(p)$ fall within $m_i$. This forms the initial 3D pseudo-label $M_i = \{ p \in \mathcal{P} \mid \pi(p) \in m_i \}$.
To mitigate noise from incorrect projections, we refine $M_i$ using a connectivity-based clustering algorithm, retaining only the largest cluster as the final pseudo-label $\tilde{M}_i$. Each point $p \in \tilde{M}_i$ inherits the instance attributes: the pseudo-label confidence is set to $S(p) = S_i$, the semantic prior to ${P}^{\text{2D}}(p) = {P}^{\text{2D}}_i$.

\noindent\textbf{Cross-View Instance Merging.}
To handle cases where a real-world object is detected in multiple views, we introduce a Cross-View Instance Merging (CVIM) module. 
For any two pseudo-labels $\tilde{M}_i$ and $\tilde{M}_j$ from different views, we compute their 3D intersection-over-union (IoU). 
If the IoU exceeds a predefined threshold, they are merged into a single instance ($\tilde{M}_{\text{merged}} = \tilde{M}_i \cup \tilde{M}_j$) as shown in Fig~\ref{fig:CVIM}, ensuring a consistent representation for each object.

\subsection{Temporal-Based Pseudo-Label Refinement}
\label{sec:Temporal-based_label_updating}

Using only a single frame for pseudo-label generation is susceptible to occlusion and sensor noise. To improve label quality, we propose two temporal refinement strategies that exploit cross-frame information from both online and offline perspectives. The offline strategy enhances the initial pseudo-labels by incorporating VFM semantic priors. The online strategy is integrated into the training stage, which refines the labels using point-wise prediction from a teacher network. Both strategies are built upon the VSV algorithm~\ref{alg:tsu}, enforcing temporal consistency across frames.

\subsubsection{Offline Refinement with VFM Priors.}
As a pre-processing step, we refine the initial pseudo-labels by reducing the noise in VFM predictions through temporal aggregation.
Specifically, the point cloud from adjacent frames $\mathcal{P}_{\text{adj}}$ is aligned to the current frame's coordinate system using the associated ego-motion transformations. We then apply the UPG pipeline (Section~\ref{sec:UGP}) to this aligned cloud, yielding a per-point 2D prior-based distribution $P^{\text{2D}}_{\text{adj}}$, which along with the aligned point cloud forms the voting input $(\mathcal{P}_{\text{adj}}, P^{\text{2D}}_{\text{adj}})$ for the VSV algorithm. Crucially, the VSV algorithm takes the current frame's point cloud and its initial labels as the data to be updated, and uses the information from the adjacent frames $(\mathcal{P}_{\text{adj}}, P^{\text{2D}}_{\text{adj}})$ as the voting input, yielding a higher-quality pseudo-labels for training.

\begin{algorithm}[H]
\caption{Voxel-Based Semantic Voting (VSV)}
\small
\label{alg:tsu}
\textbf{Input:}\\
Points to be updated $(\mathcal{P}, Y)$;\\
Points for voting $(\mathcal{P}', S)$, where $S \in \mathbb{R}^{|\mathcal{P}'| \times C}$ are the predicted score of $\mathcal{P}'$ and $C$ is the class number;\\
Ego-vehicle voxel $v_e$; Thresholds $T_n, T_s, D$.\\
\textbf{Output:} Updated labels $\hat{Y}$.

\SetKwFunction{FMain}{Voxel-Based Semantic Voting}
\SetKwProg{Fn}{Function}{:}{}
\Fn{\FMain}{
    \textbf{1. Voxelize Voting Data}\\
    $V \leftarrow \text{Voxelize}(\mathcal{P}', S)$ \text{ // Group points and scores into a voxel dictionary}\\
    
    \textbf{2. Build Voxel Voting Space}\\
    Initialize voxel label space $S$ with default value -1.\\
    \For{each voxel $v$ in $V$}{
        Let $n_v, \{s_i\}_{i=1}^{n_v}$ be the content of $v$.\\
        $dist \leftarrow ||v - v_e||_2$\\
        $T'_n \leftarrow (D / dist) \cdot T_n$\\
        $\bar{s}_v \leftarrow \frac{1}{n_v} \sum_{i=1}^{n_v} s_i$\\
        \If{$\max(\bar{s}_v) \ge T_s$ \textbf{and} $n_v \ge T'_n$}{
            $C[v] \leftarrow \underset{c}{\operatorname{argmax}}(\bar{s}_v)$
        }
    }
    
    \textbf{3. Update Pseudo-labels (Vectorized)}\\
    $V_{\mathcal{P}} \leftarrow \text{Voxelize}(\mathcal{P})$\\
    $\hat{Y} = C[V_{\mathcal{P}}]$\\
    $Mask = (\hat{Y} == -1)$ \text{ // unchanged voxels} \\
    $\hat{Y}[Mask] = Y[Mask]$\\
    
    \textbf{return} $\hat{Y}$\\
}
\end{algorithm}

\subsubsection{Online Refinement with Network Predictions.}
While offline refinement improves initial label quality, the static VFM priors may still contain noise. To address this, we introduce an online refinement strategy that enables the network to self-correct these labels during training. This is achieved through a teacher-student framework, where we leverage the temporally consistent predictions from an exponential moving average (EMA) updated teacher network~\cite{tarvainen2017meanteacher}. Specifically, we use the teacher to generate per-point semantic probabilities $P^{\text{3D}}_{\text{ema}}$ for the aligned adjacent point cloud. These predictions $(\mathcal{P}_{\text{adj}},,P^{\text{3D}}_{\text{ema}})$ are then fed into the VSV algorithm to update the pseudo-labels of the current frame. This online process enables the model to gradually overcome the initial VFM noise by using its own reliable predictions.

\subsection{VFMs Prior-Based Distillation Module}
Simply generating one-hot labels from VFMs is insufficient to capture the semantic distribution and feature information they provide, usually introducing noise and leading to overconfidence. To overcome this limitation, we propose a comprehensive strategy that distills VFM-based prior knowledge into the point-cloud network.

\subsubsection{Pseudo-Label Confidence Weighting.}
We posit that pseudo-labels of varying quality should not contribute equally to the training loss. This motivates a strategy to weight our base point-wise classification loss by the confidence score $S(p)$ associated with each point's pseudo-label. This weighting scheme ensures that high-confidence VFM priors have a more dominant impact on gradient updates. The weighted loss is formulated as:
\begin{equation}
\textstyle
\mathcal{L}_{\text{weighted}} = \frac{1}{|\mathcal{P}|} \sum_{p \in \mathcal{P}} S(p) \cdot \mathcal{L}_{\text{cls}}(p)
\label{eq:weighted_semantic_loss}
\end{equation}
where $\mathcal{L}_{\text{cls}}$ is Focal Loss~\cite{lin2017focal} and $|\cdot|$ denotes the cardinality of a set.

\subsubsection{Semantic Distribution Distillation.}
% To preserve the rich distributional information from the VFM, we propose to leverage its semantic priors as a soft label to supervise the 3D network using Kullback-Leibler (KL) divergence. This approach encourages the 3D network to learn the rich semantic distribution from the VFM, rather than prematurely committing to a hard-labeled category.
To distill the VFM’s rich semantic knowledge, we supervise the network using a distribution-based strategy with Kullback–Leibler (KL) divergence. Instead of a one-hot label, the supervision signal for each point $p$ is a softened probability distribution derived from VFM outputs, denoted as the teacher distribution $\hat{P}^{\text{2D}}(p)$. This distribution is obtained by applying a temperature-scaled softmax to the semantic prior $P^{\text{2D}}(p)$ provided by VFM. The 3D network prediction is similarly normalized into a student distribution $\hat{P}^{\text{3D}}(p)$. The distillation loss $\mathcal{L}_{\text{KL}}$ is defined as the average KL divergence between the teacher and student distributions:
% For each point $p$, we first transform its 2D semantic distribution $P_c^{\text{2D}}(p)$ (from Eq.~\ref{eq:point_attribute_assignment}) into a softened teacher distribution $\hat{P}_c^{\text{2D}}(p)$ using a temperature parameter $T$:
% \begin{equation}
%     \hat{P}_{c}^{\text{2D}}(p) = \frac{\exp(P_{c}^{\text{2D}}(p) / T)}{\sum_{c'=1}^C \exp(P_{c'}^{\text{2D}}(p) / T)},
% \label{eq:normalize_teacher_dist}
% \end{equation}
% Similarly, the 3D network's semantic prediction $P^{\text{3D}}(p)$ is normalized with a standard softmax to produce the student distribution $\hat{P}^{\text{3D}}(p)$. The soft-label loss minimizes the KL divergence between the point-wise 2D teacher and 3D student distributions by averaging over all points within all instances:
\begin{equation}
\textstyle
\mathcal{L}_{\text{kl}} = \frac{1}{\sum_{i=1}^N |\tilde{M}_i|} \sum_{i=1}^N \sum_{p \in \tilde{M}_i} D_{\text{KL}}(\hat{P}^{\text{2D}}(p) || \hat{P}^{\text{3D}}(p))
\label{eq:soft_loss}
\end{equation}
% where $D_{\text{KL}}(P || Q) = \sum_c P_c \log(P_c/Q_c)$ is the KL divergence.
\subsubsection{Cross-Modal Feature Distillation.}
To learn discriminative 3D instance representations, we employ a symmetric InfoNCE contrastive loss between pre-computed 2D features and online-generated 3D features on instance level. The 2D feature is prepared offline in the UPG module~\ref{sec:UGP}. For each instance $i$, we compute its 2D feature $z_i^{2D}$ by applying masked average pooling to the pixel-level embeddings from SAM's mask decoder within the predicted mask $m_i$. The corresponding 3D feature $z_i^{3D}$ is aggregated online from the 3D backbone's point features, as defined below, and then projected by an MLP $g(\cdot)$ to match the 2D feature dimension.
\begin{equation}
\textstyle
    z_i^{3D} = \frac{1}{|\tilde{M}_i|} \sum_{p \in \tilde{M}_i} f^{3D}(p)
\label{eq:3d_feature_aggregation}
\end{equation}

The cross-modal feature distillation loss consists of two symmetric components. The first term, $\mathcal{L}_{\text{2D} \to \text{3D}}$, treats 2D features as anchors to query the projected 3D features:
\begin{equation}
\textstyle
\mathcal{L}_{\text{2D} \to \text{3D}} = - \frac{1}{N} \sum_{i=1}^{N} \log \frac{\exp(\text{sim}(z_i^{2D}, g(z_i^{3D}))/\tau)}{\sum_{j=1}^{N} \exp(\text{sim}(z_i^{2D}, g(z_j^{3D}))/\tau)}
\end{equation}

The second term, $\mathcal{L}_{\text{3D} \to \text{2D}}$, is defined symmetrically by reversing the query-anchor roles:
\begin{equation}
\textstyle
\mathcal{L}_{\text{3D} \to \text{2D}} = - \frac{1}{N} \sum_{i=1}^{N} \log \frac{\exp(\text{sim}(z_i^{3D}, g(z_i^{2D}))/\tau)}{\sum_{j=1}^{N} \exp(\text{sim}(z_i^{3D}, g(z_j^{2D}))/\tau)}
\end{equation}

The final bidirectional distillation loss is defined as the average of both terms:
\begin{equation}
\textstyle
\mathcal{L}_{\text{distill}} = \frac{1}{2} \left( \mathcal{L}_{\text{2D} \to \text{3D}} + \mathcal{L}_{\text{3D} \to \text{2D}} \right)
\end{equation}

\subsection{Prototype-Based Contrastive Loss}
\label{sec:pcl}
To better learn the representations of discriminative features, we employ a prototype-based contrastive learning strategy. This involves constructing robust prototypes from high-confidence point samples and then pulling point features to their corresponding prototype. To enhance stability, we construct and utilize two distinct sets of prototypes, derived from reliable samples in both the current and adjacent frames.

\subsubsection{Reliable Points Selection.}
The foundation for prototypes is the selection of reliable predictions. For the current frame, our selection is guided by the intuition that points with higher pseudo-label confidence and higher predicted confidence are more reliable. Using this strategy, the reliable set of points for class $c$ from the current frame is defined as:
\begin{equation}
\textstyle
\hat{\mathcal{P}}_{c}^{\text{cur}} = \{p \in \mathcal{P}_t \mid C(p)=c, \; S(p)>T_{\text{conf}}, \;P^{3D}(p)>\phi \}
\label{eq:reliable_cur}
\end{equation} where $P^{3D}$ denotes the prediction of network.
For adjacent frames, a point $p$ is selected as a reliable sample for class $c$ only if the vote result generated by VSV algorithm of the voxel to which it belongs is class $c$.
\begin{equation}
\textstyle
\hat{\mathcal{P}}_c^{\text{adj}} = \{p \in \mathcal{P}_{adj} \mid Y_{\text{vote}}[\text{Voxelize}(p)] = c \}
\label{eq:reliable_adj}
\end{equation}

\subsubsection{Prototype Updating and Contrastive Loss.}
We compute two sets of class prototypes. First, for each iteration $t$, we estimate temporary prototypes by averaging the features from the student and teacher networks over their respective reliable sample sets:
\begin{equation}
\textstyle
    \hat{\mathcal{F}}_c^{\text{cur}}= \frac{1}{|\hat{\mathcal{P}}_c^{\text{cur}}|} \sum_{p \in \hat{\mathcal{P}}_c^{\text{cur}}} f^{3D}(p),
    \hat{\mathcal{F}}_c^{\text{adj}} = \frac{1}{|\hat{\mathcal{P}}_c^{\text{adj}}|} \sum_{p \in \hat{\mathcal{P}}_c^{\text{adj}}} f^{3D}_{\text{ema}}(p)
\label{eq:temp_prototype_estimation}
\end{equation}
At each training iteration, the prototype is updated using EMA, integrating its previous state with the current step's estimate:
\begin{equation}
\textstyle
    \mathcal{F}_c(t) = \theta \cdot \mathcal{F}_c(t-1) + (1-\theta) \cdot \hat{\mathcal{F}}_c
\label{eq:ema_update}
\end{equation}
where this update rule is applied to both $\mathcal{F}_c^{\text{cur}}(t)$ and $\mathcal{F}_c^{\text{adj}}(t)$, $\theta$ is the momentum hyperparameter.

We compute two contrastive losses that pull the features of foreground points $\mathcal{X} \subseteq \mathcal{P}_t$ in the current frame $t$ towards their corresponding class prototypes from both the current and adjacent frames, respectively:
\begin{equation}
\textstyle
    \mathcal{L}_{\text{cur}} = -\frac{1}{|\mathcal{X}|} \sum_{p \in \mathcal{X}} \log \frac{\exp(\text{sim}(f^{3D}(p), \mathcal{F}_{C(p)}^{\text{cur}})/\tau)}{\sum_{c'=1}^{N_c} \exp(\text{sim}(f^{3D}(p), \mathcal{F}_{c'}^{\text{cur}})/\tau)}
\label{eq:loss_cur}
\end{equation}
\begin{equation}
\textstyle
    \mathcal{L}_{\text{adj}} = -\frac{1}{|\mathcal{X}|} \sum_{p \in \mathcal{X}} \log \frac{\exp(\text{sim}(f^{3D}(p), \mathcal{F}_{C(p)}^{\text{adj}})/\tau)}{\sum_{c'=1}^{N_c} \exp(\text{sim}(f^{3D}(p), \mathcal{F}_{c'}^{\text{adj}})/\tau)}
\label{eq:loss_adj}
\end{equation}
The final prototype-based contrastive loss is the sum of these two components: $\mathcal{L}_{\text{pcl}} = \mathcal{L}_{\text{cur}} + \mathcal{L}_{\text{adj}}$.

\section{Total Loss}
We employ two prediction heads: one for semantic segmentation and another for instance segmentation. The semantic segmentation head is supervised by the proposed loss terms in VPD module. The instance segmentation head predicts the center offset per point and grouping points into instance, which is supervised by the L1 loss $L_{vote}$.
The overall loss function of the ALISE is defined as:
\begin{equation}
\textstyle
    \mathcal{L} = \alpha_1\mathcal{L}_{\text{weighted}} +
    \alpha_2\mathcal{L}_{\text{kl}} +
    \alpha_3\mathcal{L}_{\text{distill}} + \alpha_4\mathcal{L}_{\text{pcl}} + 
    \alpha_5\mathcal{L}_{\text{vote}}
\end{equation}
where $\alpha_1$, $\alpha_2$, $\alpha_3$, $\alpha_4$, $\alpha_5$ are hyperparameters to balance loss terms.

\section{Experiments}
\subsection{Waymo Open Dataset}
Following the weakly-supervised method YoCo~\cite{jiang2025you}, we conduct our experiments on version 1.4.0 of the Waymo Open Dataset (WOD)~\cite{sun2020scalability}, which includes both well-synchronized and aligned LiDAR points and images. The WOD consists of 1,150 sequences (over 200K frames), with 798 sequences for training, 202 sequences for validation, and 150 sequences for testing. For the 3D segmentation task, the dataset contains 23,691 and 5,976 frames for training and validation, respectively. We specifically focus on the vehicle, pedestrian, and cyclist categories for evaluation.

\subsection{Implementation Details}
\noindent\textbf{VFMs Setting.}
We set the box score threshold and the text score threshold of GroundingDINO both to 0.25. For the SAM model, we set the segmentation score threshold to 0.65.

\noindent\textbf{Evaluation Metric.}
We adopt the same evaluation metrics as YoCo. For 3D instance segmentation, we use average precision (AP) across different IoU thresholds to assess performance. For 3D semantic segmentation, we use mean IoU (mIoU) as the evaluation metric. Notebly, we calculate the final mIoU score by excluding the IoU of the background class, as the high background IoU can inflate the average score and obscure performance on foreground classes.

\noindent\textbf{Training Setting.}
We conduct experiments on SparseUnet~\cite{shi2020points} and Cylinder3D~\cite{zhou2021cylinder3d} backbone. SparseUnet and Cylinder3D is trained for 24 and 40 epochs respectively. All models are trained on 4 NVIDIA 3090 GPUs with a batch size of 8, using the AdamW~\cite{loshchilov2017adamw} optimizer. We set the hyperparameters $\alpha_1=100,\alpha_2=10,\alpha_3=1,\alpha_4=1,\alpha_5=1$, the prediction threshold $\phi=0.65$, the confidence threshold $T_{\text{conf}}=0.4$, the temperature scalar $\tau=0.5$ and the momentum factor $\theta=0.9$.

\begin{table*}[t]
    \centering
    \caption{Performance comparisons of 3D instance and semantic segmentation on Waymo validation dataset. \textbf{Bold} indicates optimal performance in label-free methods. $^*$ represents the pseudo-label generated by SAM using the corresponding annotation as visual prompts. $^\dag$ denotes the pseudo label generated by YoCo. UPG represents the pseudo-label generated by our UPG module. Abbreviations: vehicle (Veh.), pedestrian (Ped.), cyclist (Cyc.).}
    \vspace{-2mm}
    \adjustbox{max width=1.0\textwidth}{
    \begin{tabular}{@{}c@{\hspace{2pt}}c@{\hspace{2pt}}c@{\hspace{4pt}}cccc@{\hspace{4pt}}cccc@{}} 
    \toprule

    \multirow{2}{*}{Supervision} & \multirow{2}{*}{Annotation} & \multirow{2}{*}{Model} & \multicolumn{4}{c}{3D Instance Segmentation (AP)} & \multicolumn{4}{c}{3D Semantic Segmentation (IoU)} \\ 
    \cline{4-11}
    & & & mAP & Veh. & Ped. & Cyc. & mIoU & Veh. & Ped. & Cyc. \\ 
    \hline
    % --- Full Supervision ---
    & 3D Mask & Cylinder3D & 51.40 & 75.31 & 38.12 & 40.76 & 78.903 & 96.476 & 83.666 & 56.567 \\
    Full & 3D Mask & SparseUnet & 59.26 & 80.25 & 56.95 & 40.59 & 79.505 & 96.675 & 81.906 & 59.933 \\ 
    % & 3D Mask & PTv3 & 60.08 & 75.73 & 53.63 & 51.32 & 83.679 & 96.686 & 85.500 & 68.852 \\
    \cline{1-11}
    % --- Weak Supervision ---
    \multirow{8}{*}{Weak} & 3D Box & SparseUnet & 49.32 & 69.00 & 45.96 & 33.01 & 72.545 & 89.471 & 73.581 & 54.582 \\
    & 2D Box & SparseUnet & 35.48 & 44.54 & 36.84 & 25.08 & 63.831 & 74.102 & 72.113 & 45.278 \\
    & 2D Box$^*$ & SparseUnet & 45.12 & 64.06 & 40.06 & 31.23 & 75.571 & 93.418 & 77.982 & 55.312 \\
    & 2D Box & MWSIS & 48.42 & 61.45 & 45.23 & 38.59 & 75.898 & 90.369 & 78.996 & 58.329 \\
    \cline{2-11}
    & Click$^*$ & SparseUnet & 47.37 & 64.10 & 41.50 & 36.51 & 72.189 & 79.850 & 78.619 & 58.097 \\ 
    & Click$^\dag$ & YoCo & 55.35 & 67.69 & 55.25 & 43.12 & 74.770 & 81.136 & 81.716 & 64.459 \\
    % & Click$^\dag$ & Cylinder3D & 45.12 & 60.72 & 36.52 & 38.11 & 70.068 & 76.653 & 79.123 & 54.428 \\
    % & Click$^\dag$ & PTv3 & 46.59 & 62.59 & 38.12 & 39.06 & 72.776 & 82.109 & 77.056 & 59.163 \\
    \cline{1-11} % Line separating Weak and Unsupervised
    % --- Unsupervised Baseline ---
    % \multirow{3}{*}{Unsupervised baseline} & \multirow{3}{*}{None} & Cylinder3D & 38.14 & 54.77 & 34.59 & 25.07 & 63.383 & 83.721 & 72.548 & 33.880 \\
    % & & PTv3 & 39.81 & 59.27 & 37.90 & 22.28 & 63.870 & 86.456 & 73.470 & 31.684 \\
    \cline{2-11}
    % --- Our Method ---
    \multirow{4}{*}{Label-Free} 
     & \multirow{2}{*}{UPG (Baseline)} & SparseUnet & 39.82 & 60.69 & 37.75 & 21.04 & 63.112 & 83.126 & 73.388 & 32.823\\
     
     & & Cylinder3D & 38.14 & 54.77 & 34.59 & 25.07 & 63.383 & 83.721 & 72.548 & 33.880\\
    \cline{2-11}
    & \multirow{2}{*}{UPG (Ours)}
    % & ALISE (SparseUnet) & \textbf{50.76} & \textbf{64.53} & \textbf{49.16} & \textbf{38.58} & \textbf{71.309} & \textbf{90.006} & {77.302} & \textbf{46.620}\\
    & ALISE (SparseUnet) & \textbf{50.95} & \textbf{64.51} & \textbf{49.51} & {38.81} & \textbf{71.433} & \textbf{85.664} & \textbf{79.345} & {49.291}\\

    % & & ALISE (Cylinder3D) & 45.45 & 58.44 & 41.81 & 36.09 &  70.596 & 89.418 & \textbf{78.979} & 43.390\\
    & & ALISE (Cylinder3D) & 46.75 & 58.48 & 41.42 & \textbf{40.36} & 70.560 & 84.728 & {76.892} & \textbf{50.060}\\
    % \multirow{3}{*}{Unsupervised ALISE(Ours)} & \multirow{3}{*}{None} & Cylinder3D & 45.45 & 58.44 & 41.81 & \textbf{36.09} & \textbf{72.278} & 77.858 & \textbf{81.643} & \textbf{57.334} \\
    % & & PTv3 & 42.55 & 59.45 & 46.06 & 22.15 & 64.783 & 87.022 & 75.155 & 32.173\\
    \bottomrule  
    \end{tabular}
    }
    \label{tab:main results}
    \vspace{-2mm}
\end{table*}
\subsection{Results on the Waymo Open Dataset}
We compare ALISE with other weakly supervised and fully supervised methods for 3D instance and semantic segmentation. For fair comparison, we use SparseUnet as our primary network backbone, which is consistent with most baseline methods. The comprehensive results are presented in Table~\ref{tab:main results}.

For 3D instance segmentation, our label-free framework ALISE demonstrates highly competitive performance. Notably, ALISE surpasses MWSIS, a method that relies on GT 2D box supervision by a margin of {2.53\%} in mAP. Even more remarkably, our method outperforms the 2D Box$^*$ baseline by {5.83\%} in mAP. This baseline represents an upper bound for our initial pseudo-label generation, as it uses GT 2D boxes as prompts for SAM, whereas our method uses predicted boxes. Furthermore, when compared to methods utilizing BEV click annotations, our approach outperforms the SparseUnet (Click$^*$) baseline by a significant {3.58\%} in mAP. Meanwhile, our method achieves mAP improvements of 11.13\% and 8.61\%, and mIoU improvements of 8.321\% and 7.177\% on SparseUNet and Cylinder3D, respectively. While the weakly-supervised method YoCo still holds the top performance, our ALISE closes a substantial portion of the performance gap without requiring any manual labeling effort. This trade-off is highly acceptable considering the complete elimination of annotation costs.

\subsection{Results on the nuScenes Dataset}
We evaluate ALISE on the nuScenes dataset, comparing it against other methods using SparseUnet as the common backbone. For this evaluation, we focus on three main classes (vehicle, pedestrian, bicycle), merging categories such as bus, car, truck, construction vehicle, and trailer into a single unified {Vehicle} class. As presented in Table~\ref{tab:main_results_nus}, our method achieves a significant improvement over the baseline trained on initial UPG pseudo-labels and surpasses the click-supervised approach. However, a noticeable performance gap to the fully-supervised counterpart remains. We attribute this primarily to the inherent sparsity of the nuScenes dataset, which degrades the quality of the generated pseudo-labels compared to denser Waymo Open Dataset.

\begin{table*}[t]
    \centering
    \caption{Performance comparisons of 3D instance and semantic segmentation on nuScenes validation dataset.}
    \vspace{-2mm}
    \adjustbox{max width=1.0\textwidth}{
    \begin{tabular}{@{}c@{\hspace{2pt}}c@{\hspace{2pt}}c@{\hspace{4pt}}cccc@{\hspace{4pt}}cccc@{}} 
    \toprule
    \multirow{2}{*}{Supervision} & \multirow{2}{*}{Annotation} & \multirow{2}{*}{Model} & \multicolumn{4}{c}{3D Instance Segmentation (AP)} & \multicolumn{4}{c}{3D Semantic Segmentation (IoU)} \\ 
    \cline{4-11}
    & & & mAP & Veh. & Ped. & Bic. & mIoU & Veh. & Ped. & Bic. \\ 
    \hline
    % --- Full Supervision ---
    Full & 3D Mask & SparseUnet & 63.43 & 84.88 & 75.80 & 29.61 & 65.403 & 89.704 & 68.724 & 37.780 \\ 
    \hline
    % --- Weak Supervision (Simplified) ---
    Weak & Click$^*$ & SparseUnet & 37.22 & 63.54 & 40.01 & 8.11 & 40.399 & 57.698 & 50.434 & 13.066 \\ 
    \hline
    % --- Unsupervised / Label-Free ---
    \multirow{2}{*}{Label-Free} & UPG & SparseUnet & 38.97 & 63.67 & 46.53 & 6.70 & 44.983 & 66.614 & 55.097 & 13.238 \\
    & UPG & ALISE(Ours) & \textbf{45.98} & \textbf{66.33} & \textbf{55.83} & \textbf{15.78} & \textbf{50.965} & \textbf{75.962} & \textbf{62.531} & \textbf{14.401}\\
    \bottomrule  
    \end{tabular}
    }
    \label{tab:main_results_nus} % I suggest using a new label for the simplified table
    \vspace{-4mm}
\end{table*}

% \begin{table*}[t]
%     \centering
%     \caption{Performance comparisons of 3D instance and semantic segmentation on nuScenes validation dataset.}
%     \vspace{-2mm}
%     \begin{tabular}{@{}c@{\hspace{2pt}}c@{\hspace{2pt}}c@{\hspace{4pt}}cccc@{\hspace{4pt}}cccc@{}} 
%     \toprule
%     \multirow{2}{*}{Supervision} & \multirow{2}{*}{Annotation} & \multirow{2}{*}{Model} & \multicolumn{4}{c}{3D Instance Segmentation (AP)} & \multicolumn{4}{c}{3D Semantic Segmentation (IoU)} \\ 
%     \cline{4-11}
%     & & & mAP & Veh. & Ped. & Bik. & mIoU & Veh. & Ped. & Bic. \\ 
%     \hline
%     % --- Full Supervision ---
%     Full & 3D Mask & SparseUnet & 63.43 & 84.88 & 75.80 & 29.61 & 65.403 & 89.704 & 68.724 & 37.780 \\ 
%     \hline
%     % --- Weak Supervision (Simplified) ---
%     Weak & Click$^*$ & SparseUnet & 37.22 & 63.54 & 40.01 & 8.11 & 40.399 & 57.698 & 50.434 & 13.066 \\ 
%     \hline
%     % --- Unsupervised / Label-Free ---
%     \multirow{2}{*}{Label-Free} & UPG & SparseUnet & 38.97 & 63.67 & 46.53 & 6.7 & 44.983 & 66.614 & 55.097 & 13.238 \\
%      & UPG & ALISE(Ours) & \textbf{45.98} & \textbf{66.33} & \textbf{55.83} & \textbf{15.78} & \textbf{50.965} & \textbf{75.962} & \textbf{62.531} & \textbf{14.401}\\
%     \bottomrule  
%     \end{tabular}
%     \label{tab:main_results_nus} % I suggest using a new label for the simplified table
%     \vspace{-4mm}
% \end{table*}
\subsection{Ablation Study and Analysis}

\noindent\textbf{Effect of all modules.}
Table~\ref{main_ablation} presents our ablation study, demonstrating that each module progressively contributes to the final performance. Starting from the baseline, OFR provides a significant initial boost in mIoU. The subsequent inclusion of the VPD and ONR modules further enhances both metrics, with ONR yielding a particularly strong gain in mAP. Finally, integrating the PCL module achieves our best results, confirming the synergistic effect of all components.

\begin{table}[ht]
    \centering
    \caption{All modules ablation}
    \begin{tabular}{cccccc} 
            \toprule
            \multicolumn{4}{c}{Module} & \multirow{2}{*}{mIoU} & \multirow{2}{*}{mAP} \\ 
            \cmidrule(lr){1-4} 
            OFR & VPD & ONR & PCL & & \\ 
            \midrule
            \mbox{-} & \mbox{-} & \mbox{-} & \mbox{-} & 63.112 & 39.82 \\ 
            \checkmark & \mbox{-} & \mbox{-} & \mbox{-} & 67.936 & 41.83 \\ 
            \checkmark & \checkmark & \mbox{-} & \mbox{-} & 69.680 & 42.97 \\ 
            \checkmark & \checkmark & \checkmark & \mbox{-} & 70.091 & 46.74 \\ 
            \midrule
            \checkmark & \checkmark & \checkmark & \checkmark & \textbf{71.433} & \textbf{50.95} \\
            \bottomrule
        \end{tabular}
        \label{main_ablation}
\end{table}

\noindent\textbf{Effect of the UPG module.}
We conduct an ablation study to validate the effectiveness of the two key components in our UPG module: cluster-based refinement and CVIM. As shown in Table~\ref{upg_ablation_simplified}, incorporating the Cluster module brings a significant performance gain over the baseline. The subsequent addition of the CVIM module further improves the results, demonstrating that both components are essential for generating high-quality pseudo-labels.
\begin{table}[ht]
    \centering
    \caption{UPG ablation}
    \begin{tabular}{cccc} 
            \toprule
            Cluster & CVIM & mIoU & mAP \\
            \midrule
            \mbox{-} & \mbox{-} & 53.968 & 23.63 \\
            \checkmark & \mbox{-} & 58.635 & 33.48 \\
            \midrule
            \checkmark & \checkmark & \textbf{63.112} & \textbf{39.82} \\
            \bottomrule
    \end{tabular}
    \label{upg_ablation_simplified}
\end{table}

\noindent\textbf{Effect of VPD module.}
Table~\ref{semantic_loss_ablation} presents an ablation study on the components of our VPD module. The results demonstrate a consistent performance gain as each loss function is incrementally added to the baseline. The full model integrating $\mathcal{L}_{\text{weighted}}$, $\mathcal{L}_{\text{soft}}$ and $\mathcal{L}_{\text{distill}}$ achieves the best results. This confirms that all components work synergistically to improve the segmentation quality.
\begin{table}[ht]
    \centering
    \caption{VPD ablation}
    \label{semantic_loss_ablation}
    \begin{tabular}{ccccc} 
            \toprule
            $\mathcal{L}_{\text{weighted}}$ & $\mathcal{L}_{\text{soft}}$ & $\mathcal{L}_{\text{feat}}$ & mIoU & mAP \\
            \midrule
            \mbox{-} & \mbox{-} & \mbox{-} & 67.936 & 41.83 \\
            \checkmark & \mbox{-} & \mbox{-} & 68.808 & 42.40 \\
            \checkmark & \checkmark & \mbox{-} & 69.331 & 42.74 \\
            \midrule
            \checkmark & \checkmark & \checkmark & \textbf{69.680} & \textbf{42.97} \\
            \bottomrule
    \end{tabular}
\end{table}

\noindent\textbf{Effect of OFR module.}
We investigate the impact of the number of adjacent frames aggregated for our offline refinement strategy. As shown in Table~\ref{tsu_frames_ablation}, our experiments indicate that utilizing 2 adjacent frames yields the optimal balance, achieving the best mIoU and mAP scores. Beyond this point, including more frames leads to a slight decrease in performance, suggesting a diminishing return and the potential introduction of noise from distant temporal frames.

\begin{table}[ht]
    \centering
    \caption{OFR frame ablation}
    \begin{tabular}{lccccc}
    \toprule
    frame      & 0      & 1      & 2       & 3      & 4      \\
    \midrule
    mIoU      & 63.112 & 66.735 & \textbf{67.936} & 66.493 & 65.974 \\
    mAP       & 39.82  & 41.33  & \textbf{41.83}  & 41.26  & 40.78  \\
    \bottomrule
    \end{tabular}
    \label{tsu_frames_ablation}
\end{table}

Tab. \ref{vote_mode_ablation} compares two voting strategies to assess their impact on segmentation results. Employing the semantic prior distribution based on VFM achieves gains of 1.52\% in mAP and 1.998\% in mIoU, compared to using the class with the highest semantic score as a one-hot label for voting. This improvement can be attributed to the ability of the semantic prior to better capture category uncertainty and preserve fine-grained contextual information during label fusion.

\begin{table}[h]
    \centering
    \caption{OFR vote mode ablation}
            \begin{tabular}{ccc} 
            \toprule
            Vote Mode & mIoU           & mAP             \\ 
            \midrule
            none      & 63.112          & 39.82           \\
            one-hot      & 65.938          & 40.31           \\
            distribution      & \textbf{67.936}          & \textbf{41.83}           \\
            \bottomrule
            \end{tabular}
    \label{vote_mode_ablation}
\end{table}

\noindent\textbf{Effect of PCL Selection Criteria.}  
The results presented in Table~\ref{momentum_pcl} show an ablation study on the momentum hyperparameter $\theta$. We conducted experiments with three different values: 0.8, 0.9, and 0.99. The optimal performance is achieved at $\theta=0.9$.

\begin{table}[ht]
    \centering
    \caption{PCL ablation}
        \begin{tabular}{ccc} 
            \toprule
            {$\theta$} & {mIoU} & {mAP} \\
            \midrule
            0.8   & 70.531 & 50.76\\
            0.9   & \textbf{71.433} & \textbf{50.95} \\
            0.99  & 70.715 & 49.90\\
            \bottomrule
        \end{tabular}
    \label{momentum_pcl}
\end{table}
% \begin{table}[t]
%     \centering
%     \caption{Ablation on the criteria for selecting reliable points from the current frame.}
%     \vspace{-2mm}
%     \label{reliability_ablation}
%     \small
%     \begin{tabular}{cccc} 
%         \toprule
%         \multicolumn{2}{c}{Selection Criteria} & \multirow{2}{*}{mAP} & \multirow{2}{*}{mIoU} \\
%         \cmidrule(lr){1-2}
%         Distance & Confidence & & \\
%         \midrule
%         \checkmark & \mbox{-} &   49.83   &   70.762     \\
%         \mbox{-} & \checkmark &   49.25   &   70.531     \\
%         \checkmark & \checkmark & \textbf{50.76} & \textbf{71.309} \\
%         \bottomrule
%     \end{tabular}
%     \vspace{-2mm}
% \end{table}

\noindent\textbf{Pseudo-Labels Generated by YOLO.}
To validate the generalization on different VFMs, we conduct experiments using pseudo-labels generated by YOLO. The results are presented in Tab \ref{tab:yolo}. When initialized with pseudo-labels from YOLO, our ALISE framework outperforms the unsupervised application of YoCo.
\begin{table*}[htbp!]
    \centering
    \caption{Performance comparison of supervision strategies on the Waymo validation datase. $^*$ represents the pseudo-label generated by SAM using the corresponding annotation as prompts. $^\dag$ denotes the pseudo label generated by YoCo. YOLO
    refers to pseudo labels derived from YOLO prediction results.}
    \label{tab:yolo}
    \begin{tabular}{ccccc}
    \toprule
    Supervision & Annotation & Model & mAP & mIoU \\
    \midrule
    Full & 3D Mask & SparseUnet & 59.26 & 79.505 \\
    \midrule
    \multirow{2}{*}{Weak} & Click$^*$ & SparseUnet & 40.19 & 67.510 \\
     & Click$^\dagger$ & YoCo & 55.35 & 74.770 \\
    \midrule
    \multirow{2}{*}{Unsupervised} & YOLO & YoCo & 45.78 & 72.182 \\
     & YOLO & ALISE & \textbf{47.17} & \textbf{73.596} \\
    \bottomrule
    \end{tabular}
    \vspace{-2mm}
\end{table*}

\noindent\textbf{Finetuning with GT Labels.}
We conduct a finetuning experiment using varying percentages of GT labels. As shown in Table~\ref{tab:finetune_with_gt}, our model serves as an strong starting point, achieving remarkable performance with minimal supervision.
When finetuning with just {1.2\%} of the GT labels, our fine-tuned model achieves an mAP of {59.36\%}, which already surpasses the fully supervised baseline (59.26\% mAP) trained from scratch with 100\% of the data.
As the percentage of labels increases, the performance continues to climb, reaching 62.03 \%mAP when fine-tuned on all GT labels, demonstrating that ALISE can efficiently leverage additional labeled data.
\begin{table}[ht]
    \centering
    \caption{Performance of our model when fine-tuned with varying percentages of GT labels, compared against a fully supervised baseline.}
    \vspace{-2mm}
    \label{tab:finetune_with_gt}
    \begin{tabular}{lc} 
        \toprule
        {Percentage of GT Labels} & {mAP} \\
        \midrule
        1.2\% & 59.36 \\
        5\%   & 60.21 \\
        10\%  & 60.67 \\
        50\%  & 61.58 \\
        100\% (Fine-Tuned) & \textbf{62.30} \\
        \midrule
        100\% (Full Supervision) & 59.26 \\
        \bottomrule
    \end{tabular}
    \vspace{-2mm}
\end{table}

\subsection{Visualization}

\begin{figure}[ht]
    \centering
    \includegraphics[width=0.5\textwidth]{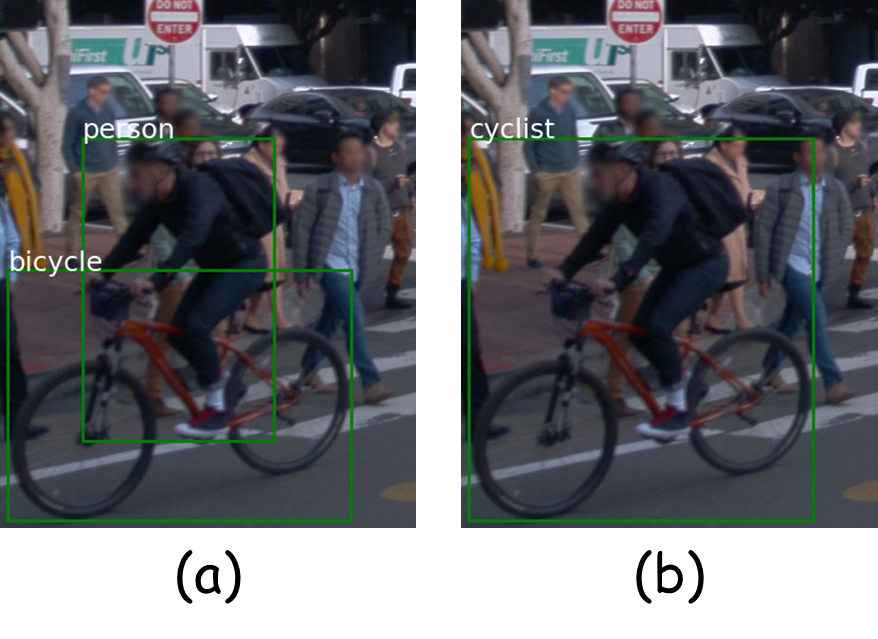}
    \caption{Visualization of rider-bicycle instance merging.}
    \label{fig:cyc}
\end{figure}

\noindent\textbf{Rider-Bicycle Instance Merging.}  
Figure~\ref{fig:cyc} shows the visualization of merging person and bicycle instances based on geometric constraints, where we evaluate the spatial relationship between detected bicycles and persons in the same frame.

\begin{figure}[ht]
    \centering
    \includegraphics[width=0.5\textwidth]{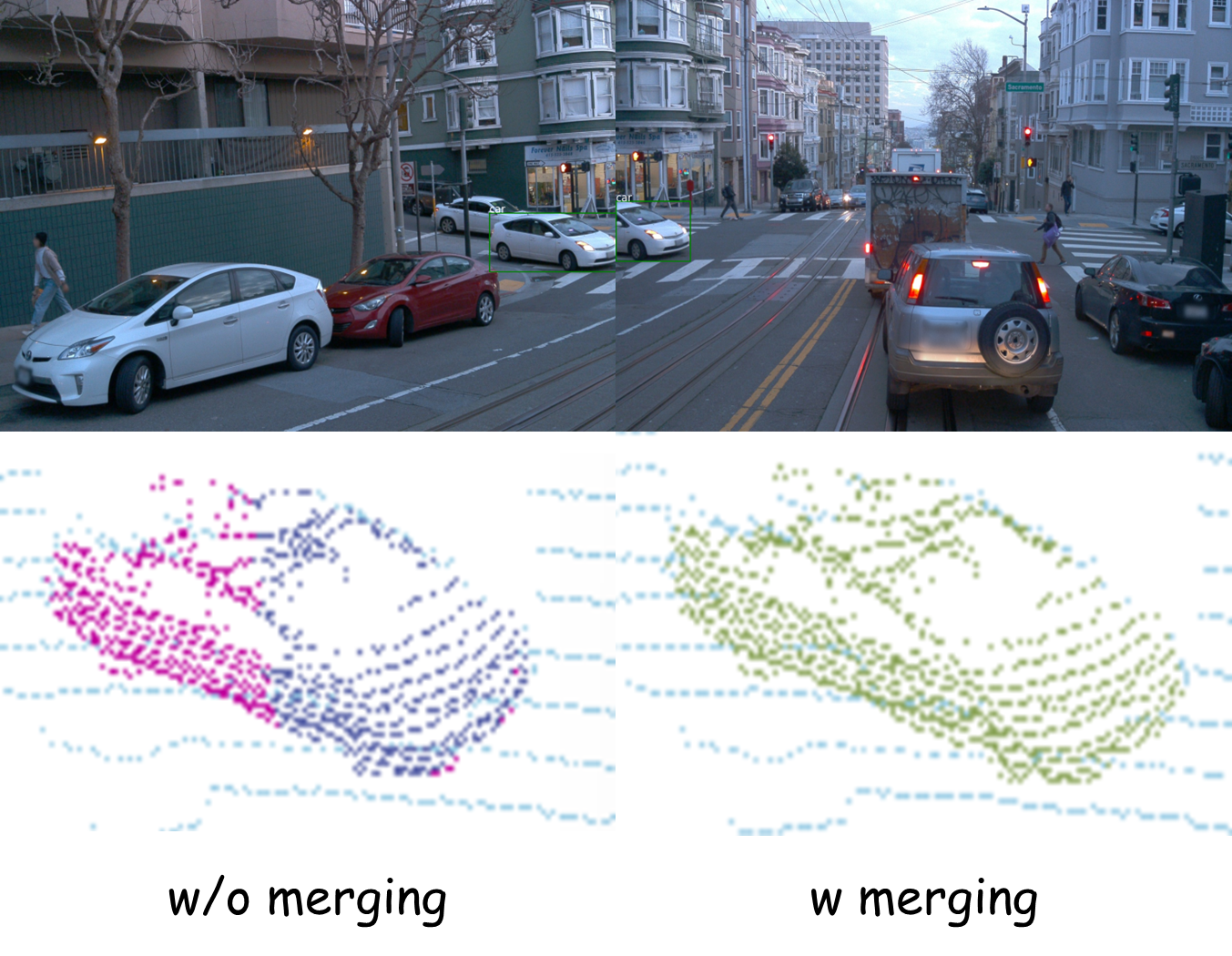}
    \caption{Visualization of cross-view instance merging (CVIM).}
    \label{fig:CVIM}
\end{figure}

\noindent\textbf{Cross-View Instance Merging.}  
The process of merging instances detected across different views or frames using 3D IoU is visualized in Figure~\ref{fig:CVIM}, which illustrates how instances from different views are merged.

\begin{figure}[ht]
    \centering
    \includegraphics[width=0.65\textwidth]{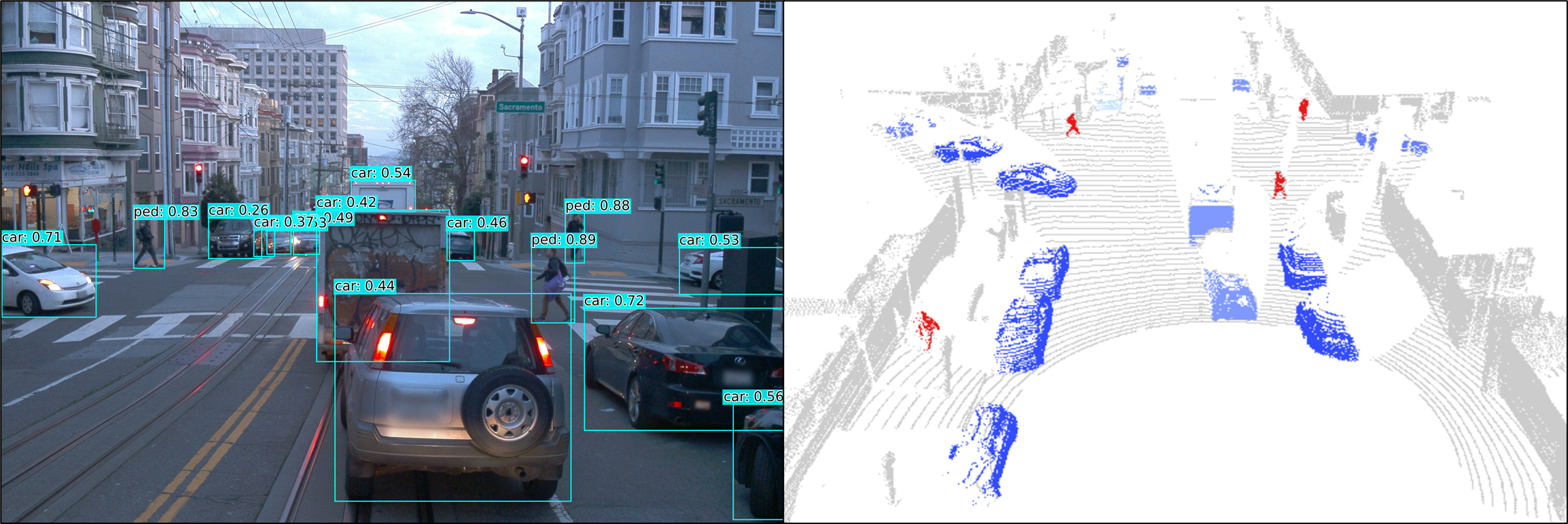}
    \caption{Visualization of pseudo-labels with confidence. Higher semantic probability corresponds to deeper color.}
    \label{fig:confidence}
\end{figure}

\noindent\textbf{Pseudo-labels with Confidence.}  
Figure~\ref{fig:confidence} illustrates the color-coding scheme used to represent the semantic class and confidence score of each detected instance. The color intensity corresponds to the confidence score, with deeper colors indicating higher detection probabilities.

\begin{figure}[ht]
    \centering
    \includegraphics[width=0.9\textwidth]{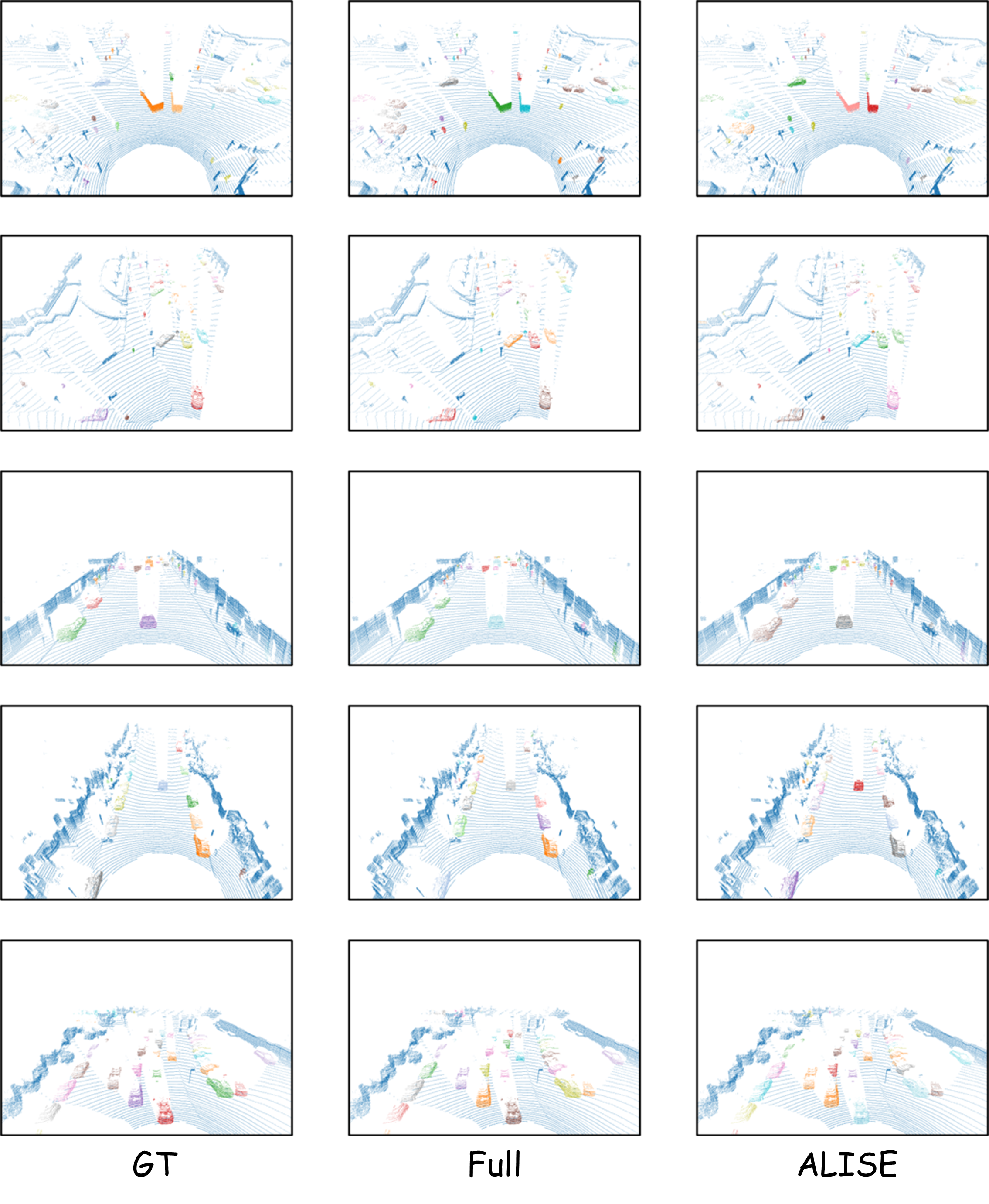}
    \caption{Visualization of instance segmentation results on Waymo Open Dataset.}
    \label{fig:vis_scene}
\end{figure}

\noindent\textbf{Instance Segmentation Results.}  
Figure~\ref{fig:vis_scene} provides a comparison of instance segmentation results on the Waymo Open Dataset, showing the performance of our unsupervised method ALISE alongside ground truth and a fully-supervised baseline.

\section{Conclusion}
In this paper, we introduced ALISE, a novel framework for annotation-free 3D instance segmentation that eliminates the dependency on manual labeling. ALISE leverages VFMs to generate initial pseudo-labels, which are then enhanced by a two-stage offline and online refinement process (OFR and ONR). For network training, we design a dual-supervision scheme with a VPD module for knowledge distillation and a PCL module for contrastive feature learning. Our experiments show that ALISE significantly improves upon unsupervised baselines and surpasses several weakly-supervised methods, demonstrating a promising direction towards automated label-free perception.
\section{Limitations}
ALISE's effectiveness is subject to limitations inherited from upstream components, whose limited understanding of specific autonomous driving scenarios may result in poor detection of certain classes like trailers and construction barriers. Furthermore, sensor calibration errors directly affect pseudo-label quality by causing spatial misalignment in the 2D-to-3D projection. These limitations are expected to be mitigated as VFMs with improved domain-specific capabilities and more advanced sensor calibration techniques become available.
\clearpage
\bibliographystyle{IEEEtran}
\bibliography{local}
\end{document}